\documentclass{article}
\usepackage{ijcai20}
\usepackage{anyfontsize}
\usepackage{times}
\usepackage{soul}
\usepackage{url}
\usepackage[hidelinks]{hyperref}
\usepackage[utf8]{inputenc}
\usepackage[small]{caption}
\usepackage{booktabs}
\usepackage{algorithmic}
\usepackage{latexsym}
\usepackage{graphicx,amsmath,amssymb}
\usepackage{fancyhdr}
\usepackage[ruled,vlined]{algorithm2e}
\include{pythonlisting}

\usepackage{microtype}

\usepackage{dsfont}
\newcommand{\RR}{\mathds{R}}
\newcommand{\OO}{\mathfrak{O}}
\newcommand{\PP}{\mathcal{P}}
\newcommand{\Lc}{\mathcal{L}}
\newcommand{\Vc}{\mathcal{V}}

\newcommand{\xv}{\mathbf{x}}
\newcommand{\yv}{\mathbf{y}}

\newcommand{\Tsf}{\mathsf{T}}
\newcommand{\Wsf}{\mathsf{W}}
\newcommand{\Xv}{\mathbf{X}}
\newcommand{\Yv}{\mathbf{Y}}
\newcommand{\inner}[2]{\langle #1, #2 \rangle}

\title{Unsupervised Multilingual Alignment using Wasserstein Barycenter}

\author{
Xin Lian$^{1,3}$
\and
Kshitij Jain$^2$\and
Jakub Truszkowski$^{2}$ \and
Pascal Poupart$^{1,2,3}$ \And
Yaoliang Yu$^{1,3}$ 
\affiliations
$^1$University of Waterloo, Waterloo, Canada\\
$^2$Borealis AI, Waterloo, Canada\\
$^3$Vector Institute, Toronto, Canada
\emails
\{x9lian, k22jain, yaoliang.yu, ppoupart\}@uwaterloo.ca,
\{jakub.truszkowski\}@borealisai.com
}

\begin{document}

\maketitle

\begin{abstract}
We study unsupervised multilingual alignment, the problem of finding word-to-word translations between multiple languages without using any parallel data. One popular strategy is to reduce multilingual alignment to the much simplified bilingual setting, by picking one of the input languages as the pivot language that we transit through. However, it is well-known that transiting through a poorly chosen pivot language (such as English) may severely degrade the translation quality, since the assumed transitive relations among all pairs of languages may not be enforced in the training process. 
Instead of going through a rather arbitrarily chosen pivot language, we propose to use the Wasserstein barycenter as a more informative ``mean'' language: it encapsulates information from all languages and minimizes all pairwise transportation costs. 
We evaluate our method on standard benchmarks and demonstrate state-of-the-art performances.

\end{abstract}

\section{Introduction}

Many natural language processing tasks, such as part-of-speech tagging, machine translation and speech recognition, rely on learning a distributed representation of words. Recent developments in computational linguistics and neural language modeling have shown that word embeddings can capture both semantic and syntactic information. This led to the development of the zero-shot learning paradigm as a way to address the manual annotation bottleneck in domains where other vector-based representations must be associated with word labels.
This is a fundamental step to make natural language processing more accessible. 
A key input for machine translation tasks consists of embedding vectors for each word.
\citet{MikolovSCCD13} were the first to release their pre-trained model and gave a distributed representation of words. After that, more software for training and using word embeddings emerged.

The rise of continuous word embedding representations has revived research on the bilingual lexicon alignment problem~\cite{Rapp:1995:IWT:981658.981709,fung1995compiling},
where the initial goal was to learn a small dictionary of a few hundred words by leveraging statistical similarities between two languages. 
\citet{MikolovLS13} formulated bilingual word embedding alignment as a quadratic optimization problem that learns an explicit \emph{linear} mapping between word embeddings, which enables us to even infer meanings of out-of-dictionary words 
\cite{zhang2016ten,Dinu2014ImprovingZL,MikolovLS13}.
\citet{Xing2015NormalizedWE} showed that restricting the linear mapping to be orthogonal further improves the result. These pioneering works required some parallel data to perform the alignment. Later on, \cite{smith2017offline,ArtetxeLA17,ArtetxeLA18b} reduced the need of supervision by exploiting common words or digits in different languages, and more recently, unsupervised methods that rely solely on monolingual data have become quite popular \cite{GouwsBC15,ZhangLLS17a,ZhangLLS17b,LampleCRDJ18,ArtetxeLA18a,DouZH18,HoshenWolf18,GraveJB19}.

Encouraged by the success on bilingual alignment, the more ambitious task that aims at \emph{simultaneously and unsupervisedly} aligning multiple languages has drawn a lot of attention recently. 

A naive approach that performs all pairwise bilingual alignment \emph{separately} would not work well, since it fails to exploit all language information, especially when there are low resource ones. 
A second approach is to align all languages to a pivot language, such as English \cite{smith2017offline},  allowing us to exploit recent progresses on bilingual alignment while still using information from all languages. 

More recently, \cite{ChenCardie18,TaitelbaumCG19a,TaitelbaumCG19b,AlauxGCJ19,WadaIM19} proposed to map all languages into the same language space and train all language pairs simultaneously. Please refer to the related work section for more details.

 In this work, we first show that the existing work on unsupervised multilingual alignment (such as \cite{AlauxGCJ19}) amounts to \emph{simultaneously} learning an arithmetic ``mean'' language from all languages and aligning all languages to the common mean language, instead of using a rather arbitrarily pre-determined input language (such as English). Then, we argue for using the (learned) Wasserstein barycenter as the pivot language as opposed to the previous arithmetic barycenter, which, unlike the Wasserstein barycenter, fails to preserve distributional properties in word embeddings.  Our approach exploits available information from all languages to enforce coherence among language spaces by enabling accurate compositions between language mappings. We conduct extensive experiments on standard publicly available benchmark datasets and demonstrate competitive performance against current state-of-the-art alternatives. 
 The source code is available at \url{https://github.com/alixxxin/multi-lang}. \footnote{Preliminary results appeared in first author thesis~\cite{Lian2020}}.
 
\section{Multilingual Lexicon Alignment}
\label{sec:bg}
In this section we set up the notations and define our main problem: the multilingual lexicon alignment problem.

Given $m$ languages $\Lc_1, \ldots, \Lc_m$, each represented by a vocabulary $\Vc_i$ consisting of $n_i$ respective words. Following \citet{MikolovLS13}, we assume a monolingual word embedding $\Xv_i = [\xv_{i,1}, \ldots, \xv_{i,n_i}]^\top \in \RR^{n_i \times d_i}$ for each language $\Lc_i$ has been trained \emph{independently} on its own data. We are interested in finding \emph{all} pairwise mappings $\Tsf_{i \to k}: \RR^{d_i} \to \RR^{d_k}$ that translate a word $\xv_{i, j_i}$ in language $\Lc_i$ to a corresponding word $\xv_{k, j_k} = \Tsf_{i \to k}(\xv_{i, j_i})$ in language $\Lc_k$. In the following, for the ease of notation, we assume w.l.o.g. that $n_i \equiv n$ and $d_i \equiv d$. Note that we do not have access to any parallel data, i.e., we are in the much more challenging unsupervised learning regime.

Our work is largely inspired by that of \citet{AlauxGCJ19}, which we review below first. Along the way we point out some crucial observations that motivated our further development. \citet{AlauxGCJ19} employ the following joint alignment approach that  minimizes the total sum of mis-alignment costs between every pair of languages: 
\begin{equation}
\label{eq:UMH}
    \!\min_{Q_i\in \OO_d, P_{ik}\in \PP_n} \sum_{i=1}^m \sum_{k=1, k\ne i}^m  \| \Xv_iQ_i - P_{ik}\Xv_k Q_k \|^2, \!
\end{equation}
where $Q_i \in \OO_d$ is a $d\times d$ orthogonal matrix and $P_{ik} \in \PP_n$ is an $n\times n$ permutation matrix\footnote{\citet{AlauxGCJ19} also introduced weights  $\alpha_{ik} > 0$ to encode the relative importance of the language pair $(i,k)$.}. Since $Q_i$ is orthogonal, this approach ensures transitivity among word embeddings: $Q_i$ maps the $i$-th word embedding space $\Xv_i$ into a common space $\Xv$, and conversely $Q_i^{-1} = Q_i^\top$ maps $\Xv$ back to $\Xv_i$. Thus, $Q_iQ_k^\top$ maps $\Xv_i$ to $\Xv_k$, and if we transit through an intermediate word embedding space $\Xv_t$, we still have the desired transitive property $Q_i Q_t^\top \cdot Q_t Q_k^\top = Q_i Q_k^\top$.

The permutation matrix $P_{ik}$ serves as an ``inferred'' correspondence between words in language $\Lc_i$ and language $\Lc_k$. Naturally, we would again expect some form of transitivity in these pairwise correspondences, i.e., $P_{ik} \cdot P_{kt} \approx P_{it}$, which, however, is not enforced in \eqref{eq:UMH}. A simple way to fix this is to decouple $P_{ik}$ into the product $P_i^\top P_k$, in the same way as how we dealt with $Q_i$. This leads to the following variant:
\begin{align}
& \underset{Q_i\in \OO_d, P_{i}\in \PP_n}{\operatorname{argmin}}
 \sum_{i=1}^m\sum_{k=1}^m \| P_i\Xv_iQ_i - P_{k}\Xv_k Q_k \|^2 \\
= & \underset{Q_i\in \OO_d, P_{i}\in \PP_n}{\operatorname{argmin}} \sum_{i=1}^m \| P_i\Xv_iQ_i - \frac{1}{m} \sum_{k=1}^m P_k \Xv_k Q_k \|^2 \label{eq:variance} \\
= & \underset{Q_i\in \OO_d, P_{i}\in \PP_n}{\operatorname{argmin}} \min_{\bar \Xv\in\RR^{n \times d}} \sum_{i=1}^m \| P_i\Xv_iQ_i - \bar \Xv \|^2 \label{eq:mean}
\end{align}
where Eq.~\ref{eq:variance} follows from the definition of variance and $\bar\Xv$ in Eq.~\ref{eq:mean}  admits the closed-form solution:
\begin{align}
\label{eq:avg}
\bar\Xv = \frac{1}{m} \sum_{k=1}^m P_k \Xv_k Q_k.
\end{align}
Thus, had we known the \emph{arithmetic} ``mean'' language $\bar\Xv$ beforehand, the joint alignment approach of \citet{AlauxGCJ19} would reduce to a separate alignment of each language $\Xv_i$ to the ``mean'' language $\bar\Xv$ that serves as the pivot. An efficient optimization strategy would then consist of alternating between separate alignment (i.e., computing $Q_i$ and $P_i$) and computing the pivot language (i.e., \eqref{eq:avg}).

\begin{figure}
    \centering
    \includegraphics[width=0.7\linewidth]{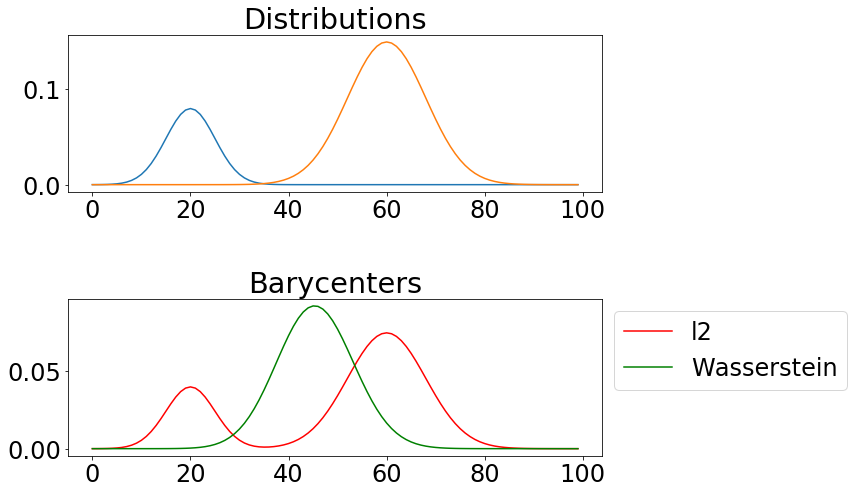}
    \caption{Comparing the Wasserstein barycenter and arithmetic mean (bottom panel) for two input distributions (top panel).}
    \label{fig:barycenter-1}
\end{figure}

We now point out two problems in the above formulation. First, a permutation assignment is a 1-1 correspondence that completely ignores polysemy in natural languages,  that is, a word in language $\Lc_i$ can correspond to multiple words in language $\Lc_k$. To address this, we propose to relax the permutation $P_i$ into a coupling matrix that allows splitting a word into different words. Second, the pivot language in \eqref{eq:avg}, being a simple arithmetic average, may be statistically very different from any of the $m$ given languages, see Figure \ref{fig:barycenter-1} and below. Besides, intuitively it is perhaps more reasonable to allow the pivot language to have a larger dictionary so that it can capture all linguistic regularities in all $m$ languages. To address this, we propose to use the Wasserstein barycenter as the pivot language.

The advantage of using Wasserstein barycenter instead of the arithmetic average is that the Wasserstein metric gives a natural geometry for probability measures supported on a geometric space. In Figure \ref{fig:barycenter-1}, we demonstrate the difference between Wasserstein Barycenter and arithmetic average of two input distributions. 

It is intuitively clear that the Wasserstein barycenter preserves the geometry of the input distributions.

\section{Our Approach}

We take a probabilistic approach, treating each language $\Lc_i$ as a probability distribution over its word embeddings:
\begin{equation}
    \pi_i = \sum_{j=1}^n p_{ij} \delta_{\xv^i_{j}} 
\end{equation}
where $p_{ij}$ is the probability of occurrence of the $j$-th word $\xv_{j}^i$ in language $\Lc_i$ (often approximated by the relative frequency of word $\xv_{j}^i$ in its training documents), and  $\delta_{\xv^i_{j}}$ is the unit mass at $\xv_{j}^i$. We project word embeddings into a common space through the orthogonal matrix $Q_i \in \OO_d$. Taking a word $\xv^i$ from each language $\Lc_i$, we associate a cost $c(Q_1\xv^1, \ldots, Q_m \xv^m) \in \RR_+$ for bundling these words in our joint translation. To allow polysemy, we find a joint  distribution $\pi$ with fixed marginals $\pi_i$ so that the average cost
\begin{align}
\label{eq:mOT}
\int c(Q_1 \xv^1, \ldots, Q_m \xv^m) ~ \mathrm{d} \pi(\xv^1, \ldots, \xv^m)
\end{align}
is minimized.
If we fix $Q_i$, then the above problem is known as multi-marginal optimal transport \cite{GangboSwiech98}.

To simplify the computation, we take the pairwise approach of \citet{AlauxGCJ19}, where we set the joint cost $c$ as the total sum of all pairwise costs:
\begin{align}
\label{eq:pwd}
\textstyle
c(\xv^1, \ldots, \xv^m) = \sum_{i, k} \|\xv^i - \xv^j \|^2.
\end{align}
Interestingly, with this choice, we can significantly simplify the numerical computation of the multi-marginal optimal transport. 

We recall the definition of Wasserstein barycenter $\nu$ of $m$ given probability distributions $\pi_1, \ldots, \pi_m$:
\begin{align}
\label{eq:wb}
\nu = \arg\min_{\mu} ~ \sum_{i=1}^m \lambda_i \cdot \Wsf_2^2(\pi_i, \mu),
\end{align}
where $\lambda \geq 0$ are the weights, and the (squared) Wasserstein distance $\Wsf_2^2$ is given as:
\begin{align}
\!\! \! \Wsf_2^2(\pi_i, \mu) \!=\!\! \min_{\Pi_i \in \Gamma(\pi_i, \mu)} \int \!\!\! \|\xv - \yv\|^2 \mathrm{d} \Pi_i(\xv, \yv). \!\!
\end{align}
The notation $\Gamma(\pi_i,\mu)$ denotes all joint probability distributions (i.e. couplings) $\Pi_i$ with (fixed) marginal distributions $\pi_i$ and $\mu$.
As proven by \citet{agueh2011barycenter}, with the pairwise distance \eqref{eq:pwd}, the multi-marginal problem in \eqref{eq:mOT} and the barycenter problem in  \eqref{eq:wb} are formally equivalent. Hence, from now on we will focus on the latter since efficient computational algorithms for it exist. We use the push-forward notation $(Q_i)_\# \pi_i$ to denote the distribution of $Q_i\xv^i$ when $\xv^i$ follows the distribution $\pi_i$. Thus, we can write our approach succinctly as:
\begin{align}
\label{eq:WBA}
\min_{\mu} \min_{Q_i\in \OO_d} ~ \sum_{i=1}^m \lambda_i \cdot \Wsf_2^2[(Q_i)_{\#} \pi_i , \mu],
\end{align}
where the barycenter $\mu$ serves as the pivot language in some common word embedding space. Unlike the arithmetic average in \eqref{eq:avg}, the Wasserstein barycenter can have a much larger support (dictionary size) than the $m$ given language distributions.

We can again apply the alternating minimization strategy to solve \eqref{eq:WBA}: fixing all orthogonal matrices $Q_i$, we find the Wasserstein barycenter using an existing algorithm of \cite{cuturi2014fast} or \cite{stochastic-barycenter}; fixing the Wasserstein barycenter $\mu$, we solve each orthogonal matrix $Q_i$ separately:
\begin{align}
\min_{Q_i \in \OO_d} \min_{\Pi_i \in \Gamma(\pi_i, \mu)} \int \|Q_i \xv - \yv\|^2 \mathrm{d} \Pi_i(\xv, \yv).\!
\end{align}
For fixed coupling $\Pi_i \in \RR^{n \times s}$, where $s$ is the dictionary size for the barycenter $\mu$, the integral can be simplified as:
\begin{align}
\!\!\sum\nolimits_{j l} (\Pi_i)_{jl} \|Q_i \xv_{j}^i - \yv_l\|^2 \!\equiv\! - \inner{\Xv_i^\top \Pi_i \Yv}{Q_i}.\!
\end{align}
Thus, using the well-known theorem of \citet{Schonemann1966}, $Q_i$ is given by the closed-form solution $U_iV_i^\top$, where $U_i\Sigma_iV_i^\top = \Xv_i^\top \Pi_i \Yv$ is the singular value decomposition. Our approach is presented in Algorithm~\ref{alg:WB}.

\begin{algorithm}[t]

	\caption{Barycenter Alignment}
	\KwIn{\text{Language distribution} $L_i=(X_i, p_i)_{i=1}^m, p$} 
	\KwOut{Translation for $L_k$ and $L_m$} 
	
	\For{$i = 1, \ldots, m$}{
		$\Xv_i \gets \Xv_i - \mathsf{mean}(\Xv_i)$ 
		
		$\{C_i\} \gets cosine\_dist(\Xv_{i,j}, \Xv_{i,k}) \forall {j,k}$
		
		$\Pi_i \gets \mathsf{GW}(C_i, C_1, p_i, p_1)$ 
		
		$U\Sigma V^\top \gets \mathsf{SVD}(\Xv_i^\top \Pi_i \Xv_1)$ 
		
		$Q_i \gets UV^\top$ 
		
		$\Xv_i \gets \Xv_i Q_i$ 
	}
	\While{not converged}{
		$\nu \gets \mathsf{WB}(\pi_1,\cdots, \pi_m; \lambda_1, \cdots, \lambda_m)$
		
		\For{$i = 1, \ldots, m$}{
			$\Pi_i \gets \mathsf{OT}(\pi_i, \nu)$ 
			
			$U\Sigma V^\top \gets \mathsf{SVD}(\Xv_i^\top \Pi_i \Yv)$
			
			$Q_i \gets UV^\top$
			
			$\Xv_i \gets \Xv_iQ_i$\ ;
			
		}
	}
	\Return $(\Pi_1, \ldots, \Pi_m, Q_1, \ldots, Q_m)$ 
	\label{alg:WB}
\end{algorithm}
\section{Experiments}
\label{sec:experiments}
We evaluate our algorithm on two standard publicly available datasets: MUSE \cite{LampleCRDJ18} and XLING \cite{glavas-etal-2019-properly}.  The MUSE benchmark is a high-quality dictionary containing up to 100k pairs of words and has now become a standard benchmark for cross-lingual alignment tasks~\cite{LampleCRDJ18}. On this dataset, we conducted an experiment with 6 European languages: English, French, Spanish, Italian, Portuguese, and German. The MUSE dataset contains a direct translation for any pair of languages in this set.
We also conducted an experiment with the XLING dataset with a more diverse set of languages: Croatian (HR), English (EN), Finnish (FI), French (FR), German (DE), Italian (IT), Russian (RU), and Turkish (TR). In this set of languages, we have languages coming from three different Indo-European branches, as well as two non-Indo-European languages (FI from Uralic and TR from the Turkic family)~\cite{glavas-etal-2019-properly}.

\subsection{Implementation Details} 
To speed up the computation, we took a similar approach as~\citet{AlauxGCJ19} and initialized space alignment matrices with the Gromov-Wasserstein approach \cite{AlvarezMelisJaakkola18} applied to the first 5k vectors (\citet{AlauxGCJ19} used the first 2k vectors) and with regularization parameter $\epsilon$ of $5e^{-5}$. 
After the initialization, we use the space alignment matrices to map all languages into the language space of the first language. Multiplying all language embedding vectors with the corresponding space alignment matrix, we realign all languages into a common language space.
In the common space, we compute the Wasserstein barycenter of all projected language distributions. The support locations for the barycenter are initialized with random samples from a standard normal distribution.

The next step is to compute the optimal transport plans from the barycenter distribution to all language distributions. After obtaining optimal transport plans $T_i$ from the barycenter to every language $\mathcal{L}_i$, we can imply translations from $\mathcal{L}_i$ to $\mathcal{L}_j$ from the coupling $T_i T_j^\top$.
The coupling is not necessarily a permutation matrix, and indicates the probability with which a word corresponds to another. 
Method and code for computing accuracies of bilingual translation pairs are borrowed from \citet{AlvarezMelisJaakkola18}.

\subsection{Baselines} We compare the results of our method on MUSE with the following methods: 1) Procrustes Matching with RSLS as similarity function to imply translation pairs \cite{LampleCRDJ18};
2) the state-of-the-art bilingual alignment method, Gromov-Wasserstein alignment (GW)~\cite{AlvarezMelisJaakkola18}; 3) the state-of-the-art multilingual alignment method (UMH)~\cite{AlauxGCJ19}; 4) bilingual alignment with multilingual auxiliary information (MPPA)~\cite{TaitelbaumCG19a}; and 5) unsupervised multilingual word embeddings trained with multilingual adversarial training~\cite{ChenCardie18}.

We compare the results of our method on XLING dataset with Ranking-Based Optimization (RCSLS)~\cite{JoulinBMJG18},  solution to the Procrustes Problem (PROC)~\cite{ArtetxeLA18a,LampleCRDJ18,glavas-etal-2019-properly}, Gromov-Wasserstein alignment (GW)~\cite{AlvarezMelisJaakkola18}, and VECMAP~\cite{ArtetxeLA18a}. RCLS and PROC are supervised methods, while GW and VECMAP are both unsupervised methods. 

The translation accuracies for Gromov-Wasserstein are computed using the source code released by~\cite{AlvarezMelisJaakkola18}.
For the multilingual alignment method (UMH)~\cite{AlauxGCJ19}, and the two multilingual adversarial methods~\cite{ChenCardie18}, ~\cite{TaitelbaumCG19a},
we directly compare our accuracies to previous methods as reported in ~\cite{glavas-etal-2019-properly}.

\subsection{Results} Table~\ref{table:experiment1} depicts precision@1 results for all bilingual tasks on the MUSE benchmark~\cite{LampleCRDJ18}. For most language pairs, our method Barycenter Alignment (BA) outperforms all current unsupervised methods. Our barycenter approach infers a ``potential universal language" from input languages. Transiting through that universal language, we infer translation for all pairs of languages. From the experimental results in Table~\ref{table:experiment1}, we can see that our approach is clearly at an advantage and it benefits from using the information from all languages. Our method achieves statistically significant improvement for 22 out of 30 language pairs ($p \leq 0.05$, McNemar's test, one-sided).

Table~\ref{table:experiment2} shows mean average precision (MAP) for 10 bilingual tasks on the XLING dataset~\cite{glavas-etal-2019-properly}.

 In Table~\ref{table:translation-prediction}, we show several German to English translations and compare the results to Gromov-Wasserstein direct bilingual alignment. Our method is capable of incorporating both the semantic and syntactic information of one word.
For example, the top ten predicted English translations for the German word \textit{München}, are ``Cambridge, Oxford, Munich, London, Birmingham, Bristol, Edinburgh, Dublin, Hampshire, Baltimore". In this case, we hit the English translation Munich. What's more important in this example is that all predicted English words are the name of some city. Therefore, our method is capable of implying \textit{München} is a city name.
Another example is the German word \textit{sollte}, which means ``should" in English.
The top five words predicted for \textit{sollte} are syntactically correct - \textit{would}, \textit{could}, \textit{will}, \textit{should} and \textit{might} are all modal verbs. 
The last three examples show polysemous words, and in all these cases our method performs better than the Gromov-Wasserstein. For German word \textit{erschienen}, our algorithm predicts all three words {\it released}, {\it appeared}, and {\it published} in the top ten translations as compared to Gromov-Wasserstein which only predicts {\it published }. 


\subsection{Ablation Study} 
\begin{figure}[ht]
    \centering
    \includegraphics[width=\linewidth]{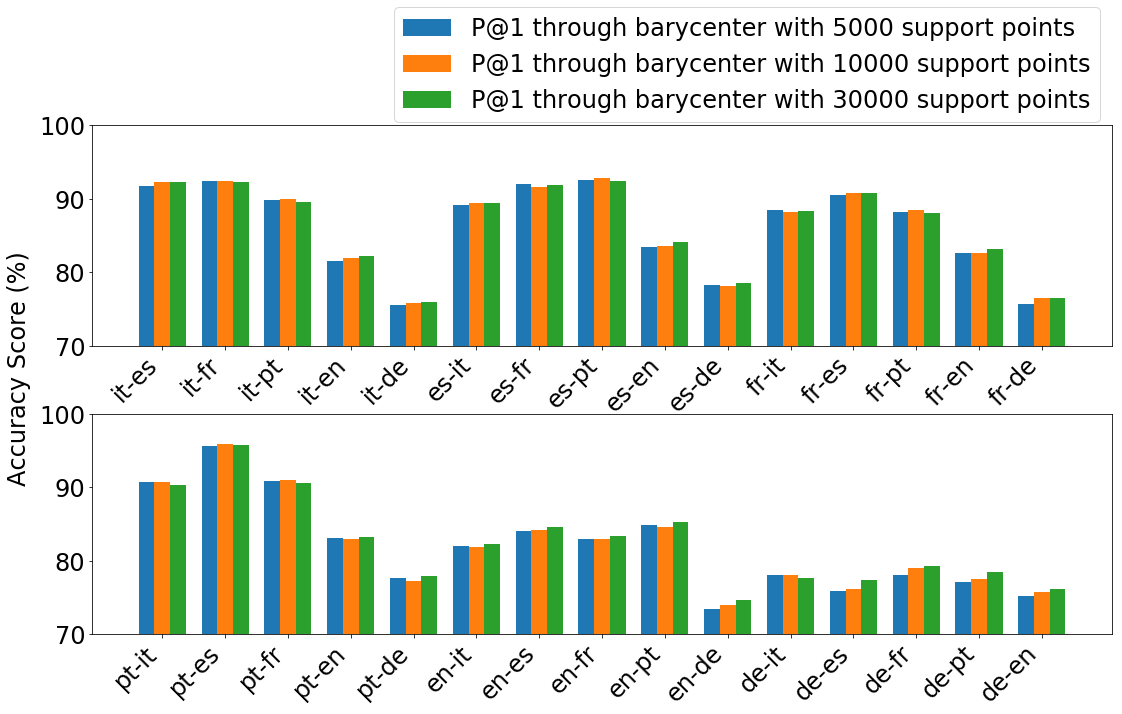}
    \caption{Accuracies for language pairs using different numbers of support locations for the barycenter. In our experimental setup, we have 5000 words in each language.
    }
\label{fig:number-of-support}
\end{figure}

In this section, we show the impact of some of our design choices and hyperparameters.
One of the parameters is the number of support locations. In theory, the optimal barycenter distribution could have as many support locations as the sum of the total number of support locations for all input distributions.
In Figure~\ref{fig:number-of-support}, we show the impact on translation performance when we have a different number of support locations. Let $n_j$ be the number of words we have in language $L_j$.
We picked the three most representative cases: the average number of words $ = \sum_{j=1}^m n_j / m$, twice the average number of words $ = 2\sum_{j=1}^m n_j / m$, and the total number of words $= \sum_{j=1}^m n_j$.
As we increase the number of support locations for the barycenter distribution, we can see in Figure~\ref{fig:number-of-support} that the performance for language translation improves. However, when we increase the number of support locations for the barycenter, the algorithm becomes costly.
Therefore, in an effort to balance accuracy and computational complexity, we decided to use 10000 support locations (twice the average number of words). 

\begin{figure}
    \centering
    \includegraphics[width=\linewidth]{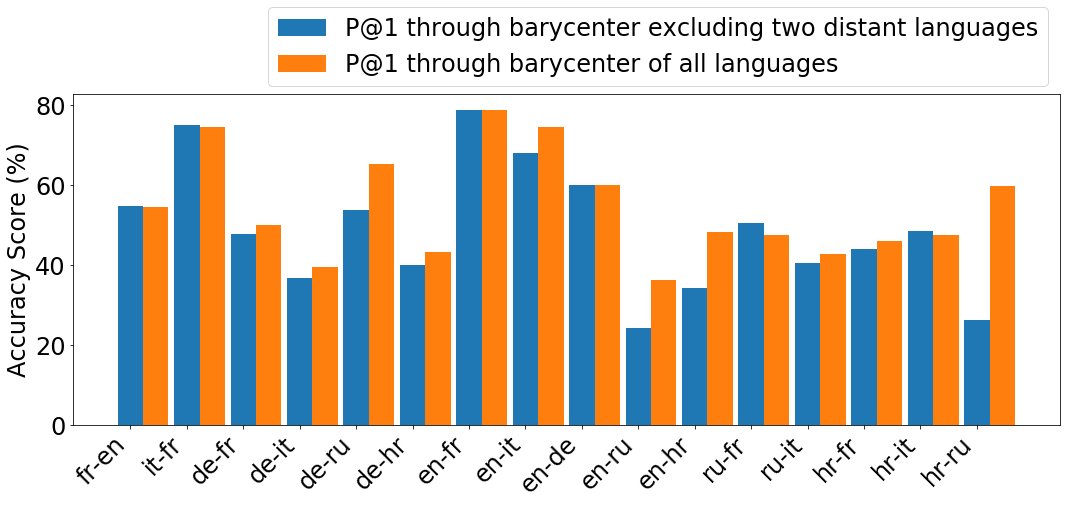}
    \caption{This graph shows the accuracy of bilingual translation pairs. The red bar indicate translation accuracy using the barycenter of all languages (HR, EN, FI, FR, DE, RU, IT, TR), while the blue bar correspond to the barycenter of (HR, EN, FR, DE, IT, RU).}
    \label{fig:compare}
\end{figure}

We also conducted a set of experiments to determine whether the inclusion of distant languages increases bilingual translation accuracy. Excluding two non-Indo-European languages Finnish and Turkish, we calculated the barycenter of 
Croatian (HR), English (EN), French (FR), German (DE), and Italian (IT).
Figure~\ref{fig:compare} contains results for common bilingual pairs. The red bar show the bilingual translation accuracy when translating through the barycenter for all languages including Finnish and Turkish, whereas blue bar indicate the accuracy of translations that use the barycenter of the five Indo-European languages.

\begin{table*}[ht]
\centering

\vspace{-.5em}

\resizebox{\textwidth}{!}{
\begin{tabular}{c|c|c|c}
\hline \hline
German  & English &  GW Prediction & BA Prediction
\\
\hline
München & Munich & London, Dublin, Oxford, Birmingham, Wellington & Cambridge, Oxford, \textbf{Munich}, London, Birmingham \\
& & Glasgow, Edinburgh, Cambridge, Toronto, Hamilton & Rristol, Edinburgh, Dublin, Hampshire, Baltimore \\
\hline
sollte & should & would, could, might, will, needs, & would, could, will, supposed,   \\
& & supposed, put, willing, wanted, meant  & might, meant, needs, expected, able, \textbf{should} \\

\hline 
        erschienen & released & \textbf{published}, editions, publication, edition & \textbf{published}, editions, volumes, publication \\
          & appeared  & printed, volumes, compilation & \textbf{released}, titled,  printed \\
          & published & publications, releases, titled &  \textbf{appeared}, edition, compilation \\         \hline
         
          aufgenommen & admitted & \textbf{recorded}, \textbf{taken}, recording, selected & \textbf{recorded}, \textbf{taken}, recording, \textbf{admitted} \\
          & recorded & roll, placed, performing &  selected, sample, \textbf{included} \\
          & taken, included & eligible, motion, assessed &  track, featured, mixed   \\
         \hline
         
         viel & lots, lot & \textbf{much}, \textbf{lot}, little, more, less  & \textbf{much}, \textbf{lot}, little, less, too \\
          & much &  bit, too, plenty, than, better  &  more, than, bit, far, \textbf{lots}\\
         \hline
        
\hline
\end{tabular}
}
\caption{German-to-English translation prediction comparing results by 1) using GW alignment to imply direct bilingual mapping and 2) using Barycenter Alignment method described in Algorithm~\ref{alg:WB}. We show top-10 translations of both methods. Last three examples show the polysemous words and their corresponding translations. 
\label{table:translation-prediction}}
\end{table*}

\begin{table*}[ht]
\centering

\begin{tabular}{@{}l@{}*{11}{c}@{}}
    \hline
      & it-es & it-fr & it-pt & it-en & it-de & es-it & es-fr & es-pt & es-en & es-de \\
      \hline
GW & \textbf{92.63} & {91.78} & {89.47} & {80.38} & {74.03} & {89.35} & {91.78} & {92.82} & {81.52} & {75.03} \\
{GW}$^o$ & - & - & - & {75.2} & {-} & - & - & - & {80.4} & - \\
PA & {87.3} & {87.1} & {81.0} & {76.9} & {67.5} & {83.5} & {85.8} & {87.3} & {82.9} & {68.3} \\
MAT+MPPA & {87.5} & {87.7} & {81.2} & {77.7} & {67.1} & {83.7} & {85.9} & {86.8} & {83.5} & {66.5} \\
MAT+MPSR & {88.2} & {88.1} & {82.3} & {77.4} & {69.5} & {84.5} & {86.9} & {87.8} & {83.7} & {69.0} \\
UMH & {87.0} & {86.7} & {80.4} & {79.9} & {67.5} & {83.3} & {85.1} & {86.3} & \textbf{85.3} & {68.7} \\
BA & 92.32 & \textbf{92.54}$^{*}$ & \textbf{90.14} & \textbf{81.84}$^{*}$ & \textbf{75.65}$^{*}$ & \textbf{89.38} & \textbf{92.19} & \textbf{92.85} & {83.5}$^{*}$ & \textbf{78.25}$^{*}$ \\
    \hline
 & fr-it & fr-es & fr-pt & fr-en & fr-de & pt-it & pt-es & pt-fr & pt-en & pt-de \\
 \hline
GW & {88.0} & {90.3} & {87.44} & {82.2} & {74.18} & {90.62} & \textbf{96.19} & {89.9} & {81.14} & {74.83} \\
{GW}$^o$ & - & - & - & {82.1} & {-} & - & - & - & {-} & - \\
PA & {83.2} & {82.6} & {78.1} & {82.4} & {69.5} & {81.1} & {91.5} & {84.3} & {80.3} & {63.7} \\
MAT+MPPA & {83.1} & {83.6} & {78.7} & {82.2} & {69.0} & {82.6} & {92.2} & {84.6} & {80.2} & {63.7} \\
MAT+MPSR & {83.5} & {83.9} & {79.3} & {81.8} & {71.2} & {82.6} & {92.7} & {86.3} & {79.9} & {65.7} \\
UMH & {82.5} & {82.7} & {77.5} & {83.1} & {69.8} & {81.1} & {91.7} & {83.6} & {82.1} & {64.4} \\
BA  & \textbf{88.38} & \textbf{90.77}$^{*}$ & \textbf{88.22}$^{*}$ & \textbf{83.23}$^{*}$ & \textbf{76.63}$^{*}$ & \textbf{91.08} & {96.04} & \textbf{91.04}$^{*}$ & \textbf{82.91}$^{*}$ & \textbf{76.99}$^{*}$ \\

    \hline
 & en-it & en-es & en-fr & en-pt & en-de & de-it & de-es & de-fr & de-pt & de-en  & \textbf{Average}\\
 \hline
GW & {80.84} & {82.35} & {81.67} & {83.03} & {71.73} & {75.41} & {72.18} & {77.14} & {74.38} & {72.85} & {82.84}\\
{GW}$^o$ & {78.9} & {81.7} & {81.3} & {-} & {71.9} & - & - & - & {-} & {72.8} & {78.04}\\
PA & {77.3} & {81.4} & {81.1} & {79.9} & {73.5} & {69.5} & {67.7} & {73.3} & {59.1} & {72.4} & {77.98} \\
MAT+MPPA & {78.5} & {82.2} & {82.7} & {81.3} & {74.5} & {70.1} & {68.0} & {75.2} & {61.1} & {72.9}  & {78.47}\\
MAT+MPSR & {78.8} & {82.5} & {82.4} & {81.5} & {74.8} & {72.0} & {69.6} & {76.7} & {63.2} & {72.9} & {79.29}\\
UMH & {78.9} & {82.5} & {82.7} & {82.0} & \textbf{75.1} & {68.7} & {67.2} & {73.5} & {59.0} & {75.5} & {78.46}\\
BA  & \textbf{81.45}$^{*}$ & \textbf{84.26}$^{*}$ & \textbf{82.94}$^{*}$ & \textbf{84.65}$^{*}$ & {74.08}$^{*}$ & \textbf{78.09}$^{*}$ & \textbf{75.93}$^{*}$ & \textbf{78.93}$^{*}$ & \textbf{77.18}$^{*}$ & \textbf{75.85}$^{*}$ & \textbf{84.24}\\
\hline
\end{tabular}

\captionof{table}{Pairs of languages in multilingual alignment problem results for English, German, French, Spanish, Italian, and Portuguese. All reported results are precision@1 percentage. The method achieving the highest precision for each bilingual pair is highlighted in bold. Methods we are comparing to in the table are: Procrustes Matching with CSLS metric to infer translation pairs (PA) ~\protect\cite{LampleCRDJ18}; Gromov-Wasserstein alignment (GW)~\protect\cite{AlvarezMelisJaakkola18} (reproduced by us using their source code); {GW}$^o$ refers to the results reported by~\protect\citet{AlvarezMelisJaakkola18} in the paper; bilingual alignment with multilingual auxiliary information (MPPA)~\protect\cite{TaitelbaumCG19a};
Multilingual pseudo-supervised refinement method~\protect\cite{ChenCardie18}; multilingual alignment method (UMH)~\protect\cite{AlauxGCJ19}. Asterisks denote significant differences between BA and GW (McNemar's test, one-sided), the only methods for which predictions were available.}
\label{table:experiment1}

\end{table*}

\begin{table*}[ht!]
\centering

\begin{tabular}{@{}l@{}*{11}{c}@{}}
    \hline
    & en-de & it-fr & hr-ru & en-hr & de-fi & tr-fr & ru-it & fi-hr & tr-hr & tr-ru\\
    \hline
    PROC (1k)  & {0.458} & {0.615} & {0.269} & {0.225} & {0.264} & {0.215} & {0.360} & {0.187} & {0.148} & {0.168}\\
    PROC (5k) & {0.544} & {0.669} & {0.372} & {0.336} & {0.359} & {0.338} & {0.474} & {0.294} & {0.259} & {0.290}\\
    PROC-B & {0.521} & {0.665} & {0.348} & {0.296} & {0.354} & {0.305} & {0.466} & {0.263} & {0.210} & {0.230}\\
    RCSLS (1k) & {0.501} & {0.637} & {0.291} & {0.267} & {0.288} & {0.247} & {0.383} & {0.214} & {0.170} & {0.191}\\
    RCSLS (5k) & {0.580} & {0.682} & {0.404} & {0.375} & {0.395} & {0.375} & {0.491} & {0.321} & {0.285} & {0.324}\\
    VECMAP & {0.521} & {0.667} & {0.376} & {0.268} & {0.302} & {0.341} & {0.463} & {0.280} & {0.223} & {0.200} \\
    GW & {0.667} & {0.751} & \textbf{0.683} & {0.123} & {0.454} & {0.485} & {0.508} & \textbf{0.634} & \textbf{0.482} & {0.295} \\
    BA & \textbf{0.683} & \textbf{0.799} & {0.667} & \textbf{0.646} & \textbf{0.508} & \textbf{0.513} & \textbf{0.512} & {0.601} & {0.481} & \textbf{0.355}\\
\hline
\end{tabular}
\captionof{table}{Mean average precision (MAP) accuracies of several current methods on XLING dataset.}
\label{table:experiment2}
\end{table*}


\section{Related Work}

We briefly describe related work on supervised and unsupervised techniques for bilingual and multilingual alignment.

\subsection{Supervised Bilingual Alignment}

\citet{MikolovLS13} formulated the problem of aligning word embeddings as a quadratic optimization problem to find an explicit linear map $Q$ between the word embeddings $\Xv_1$ and $\Xv_2$ of two languages. 
\begin{equation}
   \min_Q ~||\Xv_1 Q - P \Xv_2||_2^2 
\end{equation}
This setting is supervised since the assignment matrix $P$ that maps words of one language to another is known. Later,~\cite{Xing2015NormalizedWE} showed that the results can be improved by restricting the linear mapping $Q$ to be orthogonal. This corresponds to Orthogonal Procrustes~\cite{Schonemann1966}.  

\subsection{Unsupervised Bilingual Alignment}

In the unsupervised setting, the assignment matrix $P$ between words is unknown, and we resort to the joint optimization:
\begin{equation}
\min_Q \min_P ~||\Xv_1 Q - P \Xv_2||_2^2.
\label{eq:bilingual-unsupervised}
\end{equation}
As a result, the optimization problem becomes non-convex and therefore more challenging. The problem can be relaxed into a (convex) semidefinite program.
This method provides high accuracy at the expense of high computation complexity. Therefore, it is not suitable for large scale problems.  Another way to solve \eqref{eq:bilingual-unsupervised} is to use Block Coordinate Relaxation, where we iteratively optimize each variable with other variables fixed.  When $Q$ is fixed, optimizing $P$ can be done with the Hungarian algorithm in $O(n^3)$ time (which is prohibitive since $n$ is the number of words). \citet{cuturi2014fast} developed an efficient approximation (complexity $O(n^2)$) achieved by adding a negative entropy regularizer.
Observing that both $P$ and $Q$ preserve the intra-language distances, \citet{AlvarezMelisJaakkola18} cast the unsupervised bilingual alignment problem as a Gromov-Wasserstein optimal transport problem, and give a solution with minimum hyper-parameter to tune.

\subsection{Multilingual Alignment}

In multilingual alignment, we seek to align multiple languages together while taking advantage of inter-dependencies to ensure consistency among them.  A common approach consists of mapping each language to a common space $X_0$ by minimizing some loss function $l$:
\begin{equation}
\label{eq:UMH-all-pairs}
    \min_{Q_i\in \OO_d, P_i\in \PP_n} \sum\nolimits_i l(X_iQ_i,P_iX_0)
\end{equation}
The common space may be a pivot language such as English~\cite{smith2017offline,LampleCRDJ18,JoulinBMJG18}.  \citet{nakashole-flauger-2017-knowledge} and \citet{AlauxGCJ19} showed that constraining coherent word alignments between triplets of nearby languages improves the quality of induced bilingual lexicons.  
\citet{ChenCardie18} extended the work of \cite{LampleCRDJ18} to the multilingual case using adversarial algorithms. \citeauthor{TaitelbaumCG19b} extended Procrustes Matching to the multi-Pairwise case \cite{TaitelbaumCG19a}, and also designed an improved representation of the source word using auxiliary languages \cite{TaitelbaumCG19b}.

\section{Conclusion}

In this paper, we discussed previous attempts to solve the multilingual alignment problem, compared similarity between the approaches and pointed out a problem with existing formulations.
Then we proposed a new method using the Wasserstein barycenter as a pivot for the multilingual alignment problem. 
At the core of our algorithm lies a new inference method based on an optimal transport plan to predict the similarity between words. 
Our barycenter can be interpreted as a virtual universal language, capturing information from all languages. 
The algorithm we proposed improves the accuracy of pairwise translations compared to the current state-of-the-art method..

\section*{Acknowledgments}

We thank the reviewers for their critical comments and we are grateful for funding support from 
NSERC and Mitacs.

\clearpage

\bibliographystyle{named}
\bibliography{acl2020,ijcai20}

\section{Appendix}

\subsection{Barycenter Convergence} 
\begin{figure*}
    \centering
    \includegraphics[width=\linewidth]{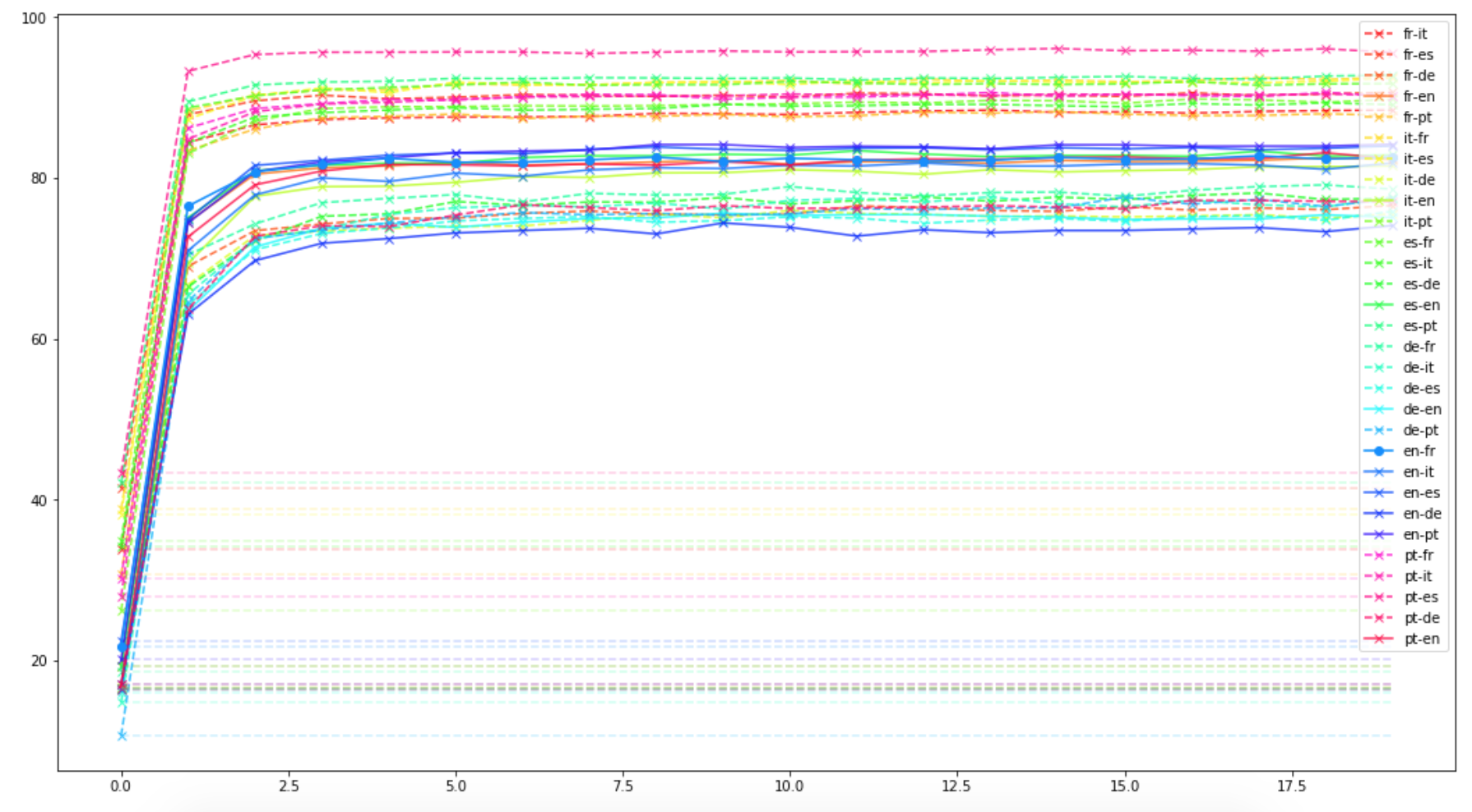}
    \caption{Translation accuracies for language pairs as a function of the number of iterations. The barycenter stabilizes after the $5$-th iteration.}
    \label{fig:convergence}
\end{figure*}

Each iteration of our barycenter algorithm optimizes the barycenter weights and then the support locations. In this section, we investigate the speed of convergence of our approach. 
In figure \ref{fig:convergence}, we plot the translation accuracy for all language pairs as a function of the number of iterations. As we can see, the accuracy stabilizes after roughly $5$ iterations.

\subsection{Hierarchical Approach} 
Training a joint barycenter for all languages captures shared information across all languages. We hypothesized that distant languages might potentially impair performance for some language pairs. To leverage existing knowledge of similarities between languages, we constructed a language tree whose topology was consistent with the widely agreed phylogeny of Indo-European languages (see e.g.~\cite{gray2003language}). For each non-leaf node, we set it the barycenter for all its children. We traverse the language tree in depth-first order and store the mappings corresponding to each edge. The translation between any two languages can be implied by traversing through the tree structure and multiplying the mappings corresponding to each edge.

Table~\ref{table:total-res} shows the results for the hierarchical barycenter. We see that the hierarchical approach yields slightly better performance for some language pairs, particularly for closely related languages such as Spanish and Portuguese or Italian and Spanish. For most language pairs, it does not improve over the weighted barycenter. More details about the hierarchical approach are available in first author's thesis~\cite{Lian2020}.

\begin{table*}[t]
\centering
\begin{tabular}{c|cc|cc|cc|cc}
\hline
& \multicolumn{2}{l}{GW benchmark} & \multicolumn{2}{l}{unweighted} &
\multicolumn{2}{l}{hierarchical} &
\multicolumn{2}{l}{weighted}
\\
& P@1 & P@10 & P@1 & P@10 & P@1 & P@10 & P@1 & P@10 \\
\hline 
it-es  &  \bf 92.63  &  98.05  &  91.52  &  97.95  &  92.49  &  \bf 98.11  &  92.32  &  98.01 \\
it-fr  &  91.78  &  98.11  &  91.27  &  97.89  &  \bf 92.61  &  \bf 98.14  &  92.54  &  \bf 98.14 \\
it-pt  &  89.47  &  97.35  &  88.22  &  97.25  &  89.89  &  \bf 97.87  &  \bf 90.14  &  97.84 \\
it-en  &  80.38  &  93.3  &  79.23  &  93.18  &  79.54  &  93.21  &  \bf 81.84  &  \bf 93.77 \\
it-de  &  74.03  &  93.66  &  74.41  &  92.96  &  73.06  &  92.26  &  \bf 75.65  &  \bf 93.82 \\
es-it  &  89.35  &  97.3  &  88.8  &  97.05  &  \bf 89.73  &  \bf 97.5  &  89.38 &  97.43 \\
es-fr  &  91.78  &  98.21  &  91.34  &  98.03  &  91.74  &  98.29  &  \bf 92.19  &  \bf 98.33 \\
es-pt  &  92.82  &  98.32  &  91.83  &  98.18  &  92.65  &  \bf 98.35  &  \bf 92.85  &  \bf 98.35 \\
es-en  &  81.52  &  94.79  &  82.43  &  94.63  &  81.63  &  94.27  &  \bf 83.5  &  \bf 95.48 \\
es-de  &  75.03  &  93.98  &  76.47  &  93.73  &  74.86  &  93.73  &  \bf 78.25  &  \bf 94.74 \\
fr-it  &  88.0  &  97.5  &  87.55  &  97.19  &  88.35  &  97.64  &  \bf 88.38  &  \bf 97.71 \\
fr-es  &  90.3  &  97.97  &  90.18  &  97.68  &  90.66  &  \bf 98.04  &  \bf 90.77  &  \bf 98.04 \\
fr-pt  &  87.44  &  96.89  &  86.7  &  96.79  &  \bf 88.35  &  \bf 97.11  &   88.22  &  97.08 \\
fr-en  &  82.2  &  94.19  &  81.26  &  94.25  &  80.89  &  94.13  &  \bf 83.23  &  \bf 94.42 \\
fr-de  &  74.18  &  92.94  &  74.07  &  92.73  &  74.44  &  92.68  &  \bf 76.63  &  \bf 93.41 \\
pt-it  &  90.62  &  97.61  &  89.36  &  97.75  &  90.59  &  \bf 98.17  &  \bf 91.08  &  97.96 \\
pt-es  &  \bf 96.19  &  99.29  &  95.36  &  99.08  &  96.04  &  99.23  &  96.04  &  \bf 99.32 \\
pt-fr  &  89.9  &  97.57  &  90.1  &  97.43  &  90.67  &  97.74  &  \bf 91.04  &  \bf 97.87 \\
pt-en  &  81.14  &  94.17  &  81.42  &  94.14  &  81.42  &  93.86  &  \bf 82.91  &  \bf 94.64 \\
pt-de  &  74.83  &  93.76  &  75.94  &  93.21  &  74.45  &  93.1  &  \bf 76.99  &  \bf 94.32 \\
en-it  &  80.84  &  93.97  &  79.88  &  93.93  &  80.25  &  93.76  &  \bf 81.45  &  \bf 94.58 \\
en-es  &  82.35  &  94.67  &  83.05  &  94.79  &  81.62  &  94.82  &  \bf 84.26  &  \bf 95.28 \\
en-fr  &  81.67  &  94.24  &  81.86  &  94.33  &  81.42  &  93.99  &  \bf 82.94  &  \bf 94.67 \\
en-pt  &  83.03  &  94.45  &  82.72  &  94.64  &  82.25  &  94.79  &  \bf 84.65  &  \bf 95.29 \\
en-de  &  71.73  &  90.48  &  72.92  &  90.76  &  71.88  &  90.42  &  \bf 74.08  &  \bf 91.46 \\
de-it  &  75.41  &  94.3  &  76.4  &  93.87  &  75.19  &  93.65  &  \bf 78.09  &  \bf 94.52 \\
de-es  &  72.18  &  92.64  &  74.21  &  92.6  &  73.58  &  92.48  &  \bf 75.93  &  \bf 93.83 \\
de-fr  &  77.14  &  93.29  &  77.93  &  93.61  &  77.14  &  93.51  &  \bf 78.93 &  \bf 93.77 \\
de-pt  &  74.38  &  93.71  &  74.99  &  93.54  &  74.22  &  93.81  &  \bf 77.18  &  \bf 94.14 \\
de-en  &  72.85  &  91.06  &  74.36  &  91.21  &  72.17  &  90.81  &  \bf 75.85  &  \bf 91.98 \\
average  &  82.84  &  95.26  &  82.86  &  95.15  &  82.79  &  95.18  &  84.24  &  95.67
  \\ \hline
\end{tabular}
\caption{Accuracy results for translation pairs between all pairs of languages for all evaluated methods. The column GW-benchmark contains results from Gromov-Wasserstein direct bilingual alignment. Unweighted is the barycenter approach without optimizing on support location weights. Hierarchical contains results from traversing through edges and infer translation mapping through hierarchical barycenters. The weighted column is what Algorithm \ref{alg:WB} returns, optimizing both on support locations and weights on the support.}
\label{table:total-res}
\end{table*}

\end{document}